\documentclass[sigconf,screen,nonacm]{acmart}

\AtBeginDocument{%
  }

\setcopyright{acmlicensed}
\copyrightyear{2025}
\acmYear{2025}
\acmConference[MM '25]{Proceedings of the 33rd ACM International Conference on Multimedia}{October 27--31, 2025}{Dublin, Ireland}
\acmBooktitle{Proceedings of the 33rd ACM International Conference on Multimedia (MM '25), October 27--31, 2025, Dublin, Ireland}\acmDOI{10.1145/3746027.3755529}
\acmISBN{979-8-4007-2035-2/2025/10}

\newcommand{\eg}{\emph{e.g.}}

\newcommand{\latentemb}{\hat{h}_k}
\newcommand{\prevlatentembset}{\{\hat{h}_i \}_{1<=i<k}}

\begin{document}

\title{StrandDesigner: Towards Practical Strand Generation with Sketch Guidance}



\author{Na Zhang}
\authornote{Both authors contributed equally to this research.}
\authornote{During internship at Tencent YouTu Lab.}
\affiliation{%
  \institution{Fudan University}
  \city{Shanghai}
  \country{China}
}
\email{nazhang23@m.fudan.edu.cn}

\author{Moran Li}
\authornotemark[1]
\affiliation{%
  \institution{Tencent YouTu Lab}
  \city{Shanghai}
  \country{China}
}
\email{moranli@tencent.com}

\author{Chengming Xu}
\authornote{Corresponding author.}
\affiliation{%
  \institution{Tencent YouTu Lab}
  \city{Shanghai}
  \country{China}
}
\email{chengmingxu@tencent.com}

\author{Han Feng, Xiaobin Hu, Jiangning Zhang,  Weijian Cao, Chengjie Wang}
\affiliation{
  \institution{Tencent YouTu Lab}
  \city{Shanghai}
  \country{China}}

\author{Yanwei Fu} 
\authornote{Prof. Yanwei Fu is with School of Data Science, Fudan University, Shanghai Innovation Institute, Institute of Trustworthy Embodied AI, Fudan University.}
\affiliation{%
  \institution{Fudan University}
  \city{Shanghai}
  \country{China}
}
\email{yanweifu@fudan.edu.cn}

\renewcommand{\shortauthors}{Zhang et al.}

\begin{abstract}
    Realistic hair strand generation is crucial for applications like computer graphics and virtual reality. While diffusion models can generate hairstyles from text or images, these inputs lack precision and user-friendliness. Instead, we propose the first sketch-based strand generation model, which offers finer control while remaining user-friendly. Our framework tackles key challenges, such as modeling complex strand interactions and diverse sketch patterns, through two main innovations: a learnable strand upsampling strategy that encodes 3D strands into multi-scale latent spaces, and a multi-scale adaptive conditioning mechanism using a transformer with diffusion heads to ensure consistency across granularity levels. Experiments on several benchmark datasets show our method outperforms existing approaches in realism and precision. Qualitative results further confirm its effectiveness. Code will be released at \href{https://github.com/fighting-Zhang/StrandDesigner}{GitHub}.

\end{abstract}



\maketitle

\section{Introduction  \label{sec:intro}}
Generating realistic hair strands is a crucial challenge in computer graphics, virtual reality, and digital content creation. High-quality strand generation can substantially elevate the visual realism of digital avatars, video game characters, and virtual humans. Recent advancements in generative models, especially diffusion models, have enabled the creation of detailed hairstyles controlled by text prompts. For example, HAAR~\cite{sklyarova2024text} proposed leveraging latent diffusion model to generate guiding strand latent features conditioned on prompts, which are upscaled to full strands using composite strategies such as nearest neighbor and bilinear interpolation.

\begin{figure}[t]
    \centering
    \includegraphics[width=\linewidth]{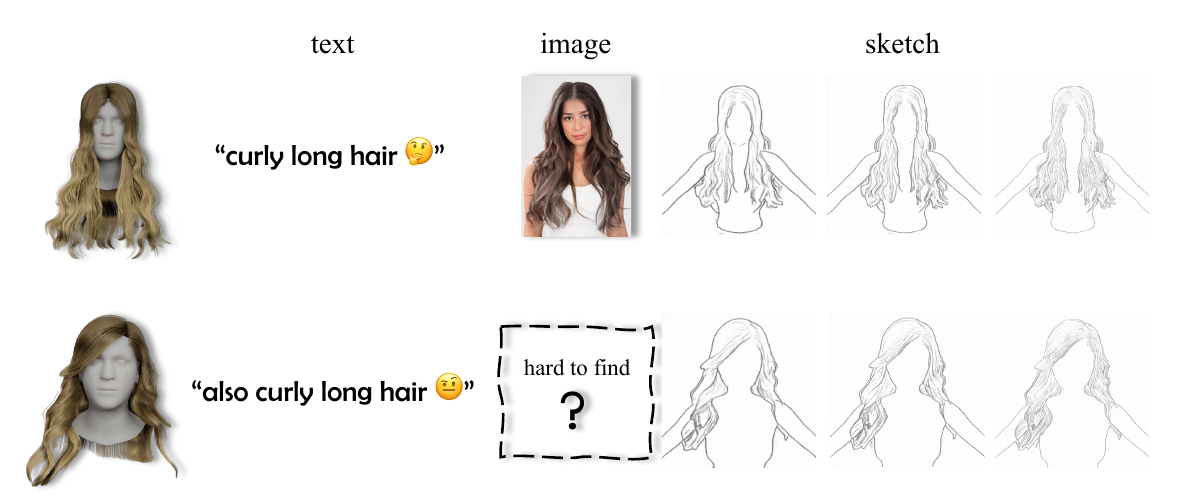}
    \caption{Text prompts often fail to describe hairstyles precisely, and finding exact reference images is challenging. By contrast, sketches are generally clearer and more flexible.}
    \label{fig:teaser}
\end{figure}

While these methods show promising potential in using visual generative models for strand generation, we argue that \textit{text or raw image prompts are not ideal for conditional strand generation}. Instead, sketch images with binarized strokes are more suitable. The key differences are illustrated in Fig.~\ref{fig:teaser}: (1) Text-based conditioning often leads to a one-to-many mapping, making precise control difficult. For example, ``curly long hair'' can correspond to many variations in curl patterns and lengths, lacking specificity for detailed strand generation. Furthermore, text descriptions are often ambiguous and may not capture the intricate details required for high-quality strand generation. (2) Image-based conditioning provides more detail but faces challenges about user-friendliness. Photographing real hairstyles is time-consuming, and static images may not fully capture hair dynamics across angles and lighting. Building a diverse dataset is also resource-intensive. (3) Sketch images offer a compelling alternative for strand generation. Sketches allow users to draw desired hairstyles with precision, capturing shape and flow more effectively than text or images. This control is valuable for customization. Moreover, sketches are inherently flexible and can be easily modified to reflect changes in hairstyle preferences or design requirements.

Based on such clues, this paper focuses on proposing the first sketch-based strand generation model. Despite these advantages, the inherent challenges posed by sketch images make it difficult to directly adapt previous methods to the new prompt format, including: (1) \textbf{Complex modeling of neighboring strands.} Sketches contain fine-grained structural information that requires sophisticated modeling techniques. Fixed upsampling strategies used by methods like HAAR, such as nearest-neighbor or bilinear interpolation, are insufficient for capturing these nuanced details, as also noted by prior works such as GroomGen~\cite{zhou2023groomgen}. These traditional methods fail to account for intricate relationships between neighboring strands, leading to suboptimal results. The complexity of hair strand interactions requires advanced modeling approaches that can accurately represent the dependencies and variations in strand configurations. (2) \textbf{Varied sketch patterns for different strand densities.} Intuitively, users’ expertise can produce sketches with divergent patterns. Professional designers may create detailed full-hairstyle sketches, while amateurs provide only sparse guiding strands. This diversity makes single-conditioning approaches ineffective, necessitating adaptive mechanisms to handle different sketch densities while maintaining realistic strand generation.

To address these challenges, we propose a novel framework incorporating a learnable strand upsampling strategy and multi-scale adaptive conditioning based on an autoregressive model. Our framework consists of two key components. First, we employ the idea of “next-scale prediction” to build a learning-to-upsample strand generation strategy. This involves encoding the 3D strands into multiple latent spaces corresponding to different guiding scales, with the smallest scale representing basic guiding strands and other scales representing residual latents involved in gradually adding more guiding strands until the full strands are built. Next, the conditional distribution of each scale's latent embeddings given previous scales is modeled by a transformer model to which diffusion heads are attached, following MAR~\cite{li2024autoregressive}. This architecture effectively utilizes continuous latent embeddings and avoids performing discrete tokenization, which is empirically proven ineffective. 

Second, we introduce a multi-scale adaptive conditioning mechanism to endow our model with the ability to handle diverse sketch patterns. This mechanism involves leveraging learnable visual tokens for sketches at each scale. For each scale, these visual tokens are fed to the pretrained DINOv2~\cite{oquab2023dinov2} to adapt the extracted features. By aligning the adapted features of sketch images with various scales with the original DINOv2 features of sketches belonging to the specific scale, the visual tokens learn to maintain consistency with the input sketch across various granularity levels. Based on the adapted sketch embeddings, we build a dual-level conditioning mechanism: local patch tokens guide fine-grained details via attention layers, while the global tokens enforce global shape consistency through direct summation.

To validate the effectiveness of our proposed method, we conduct comprehensive experiments and comparisons on the USC-HairSalon~\cite{hu2015single} and CT2Hair~\cite{shen2023ct2hair} datasets. Our method significantly outperforms other competitors across multiple metrics (e.g., Point Cloud IoU, Chamfer Distance, Hausdorff Distance, CLIP Score~\cite{hessel2021clipscore}, and LPIPS~\cite{zhang2018unreasonable}). The results demonstrate that our framework generates strands that accurately reflect input sketches and exhibit realistic interactions and patterns. 

Overall, our contributions are summarized as follows:

\begin{itemize}
\item We propose the first sketch-conditioned strand generation framework, solving both the challenges of text ambiguity and difficulty in collecting photographic data.

\item We propose the multi-scale learnable upsampling strategy as an alternative to the previously used fixed upsampling methods.

\item In order to address the intrinsic variety of sketch images, we propose the adaptive multi-scale conditioning mechanism via adaptation of pretrained DINOv2.

\end{itemize}
\section{Related Work}

\subsection{3D Hair Representation}

Early parametric methods for 3D hair representation explored various approaches. Yang et al.~\cite{xu2001v} and Wang et al.~\cite{wang2004hair} developed generalized cylinders and hierarchical cluster models, while DeepMVSHair~\cite{kuang2022deepmvshair} leveraged continuous direction fields for multi-view reconstruction. Bhokare et al.~\cite{bhokare2024real} later demonstrated real-time rendering with hair meshes. While these methods provided intuitive styling control, they struggled with complex geometric details. Recent volumetric and implicit representations have shown promising results, with MonoHair~\cite{wu2024monohair} and TECA~\cite{zhang2024teca} employing Neural Radiance Fields (NeRFs) for coarse hair geometry and UniHair~\cite{zheng2024towards} leveraging Gaussian Splatting for single-view reconstruction. However, these approaches only model the visible surface without internal structure, making them incompatible with downstream applications.
As a practical solution for high-fidelity hair modeling, strand-based representations have become prevalent in both research~\cite{shen2023ct2hair} and industry~\cite{chiang2015practical, fascione2018path}. Neural Haircut~\cite{sklyarova2023neural} and HAAR~\cite{sklyarova2024text} advanced this direction by mapping each strand on the scalp surface to a hair map through UV-space parameterization, using a strand-VAE to compress each 3D curve into a 64-D vector.
Our approach further decomposes the strand map into multiple latent spaces, enabling a coarse-to-fine generation process that progressively increases detail from guide strands to full resolution. This design naturally aligns with professional hair modeling workflows.

\subsection{3D Hair Generation}

3D hair generation has garnered growing research interest. Early attempts employed example-based methods~\cite{ren2021hair} and volumetric VAEs~\cite{saito20183d}, yet suffered from limited diversity and over-smoothing. For unconditional generation, GroomGen~\cite{zhou2023groomgen} introduced a hierarchical VAE framework to model detailed strand geometry. More recently, Perm~\cite{he2024perm} used PCA-based parameterization for frequency decomposition, while Curly-Cue~\cite{wu2024curly} proposed algorithms to model high-frequency helical structures in tightly coiled hair.

Recent works have investigated text-guided generation, with TECA~\cite{zhang2024teca} using NeRF-based representations and HAAR~\cite{sklyarova2024text} adopting UV-space latent diffusion. However, text descriptions often fail to effectively convey users' detailed hairstyle requirements. In contrast, images and sketches offer more direct and precise control over hair details. HairStep~\cite{zheng2023hairstep} leveraged intermediate representations extracted from images for single-view reconstruction, yet struggled with occluded regions. DeepSketchHair~\cite{shen2020deepsketchhair} employed GANs to generate 3D orientation fields from multi-view sketches, but exhibited limited generation quality and practical utility.
Our method enables intuitive control via sketch inputs while maintaining consistency in strand-based hairstyle generation, seamlessly integrating with physics-based rendering and simulation.

\subsection{Coarse-to-Fine Interpolation}

A complete strand-based hairstyle typically comprises tens of thousands of hair strands. In conventional workflows, artists first create guide strands to define the overall shape, then gradually increase the density to achieve full strands. 
Most hair generation methods follow this coarse-to-fine paradigm. HAAR~\cite{sklyarova2024text} first generates a low-resolution strand map via latent diffusion, then computes blending weights for nearest neighbor and bilinear interpolation employing fixed non-linear functions based on cosine similarity between adjacent strands. For highly coiled hair, Curly-Cue~\cite{wu2024curly} designs a transported displacement function for assigned guide strands, enforcing phase locking in strictly-guided regions. GroomGen~\cite{zhou2023groomgen} and Perm~\cite{he2024perm} employ neural networks to model the blending weights of neighboring guide hairs. After interpolation, GroomGen further increases density through parameter-controlled variations in the frequency domain.
Beyond hair modeling, the coarse-to-fine paradigm has demonstrated great potential in image generation. VAR~\cite{tian2024visual} employs a multi-scale autoregressive framework with next-scale prediction for computer vision. SIT~\cite{esteves2024spectral} tokenizes the image spectrum, achieving coarse reconstruction with only a small number of tokens, enabling autoregressive models for image upsampling.
Our method adapts the next-scale prediction approach to 3D hair generation, enabling learning an upsampling paradigm for each subsequent scale based on preceding scale features. Leveraging our residual design, the latent map of each generated scale can be decoded into a hairstyle with increasingly fine details, offering users flexible control over the generation process.

\begin{figure*}[t]
  \centering
  \includegraphics[width=0.9\linewidth]{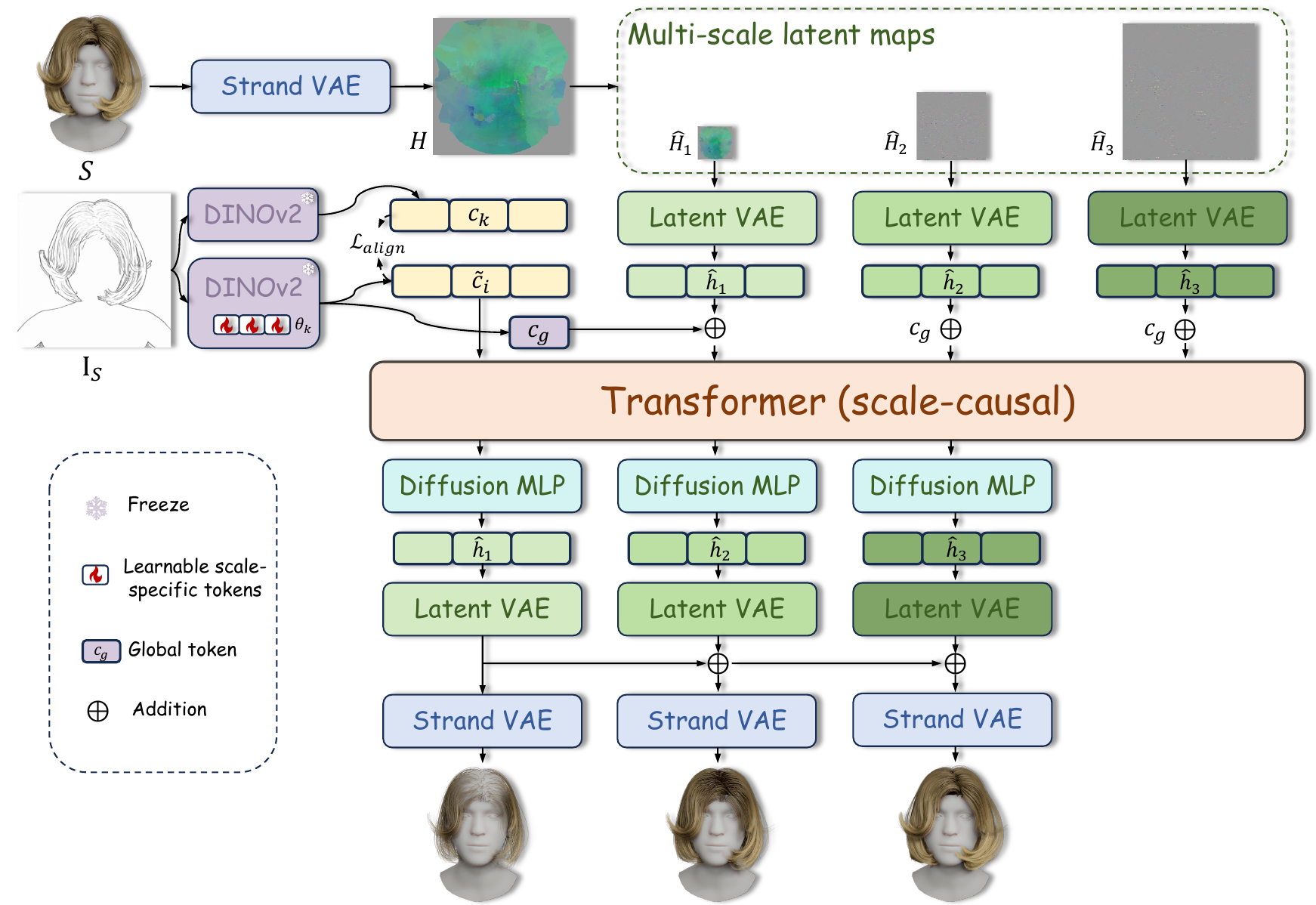}
  \caption{Overview of our proposed framework employing a learnable multi-scale upsampling strategy via next-scale prediction and an adaptive conditioning mechanism with learnable visual tokens for sketch-based strand generation.}
  \label{fig:strand_model}
\end{figure*}

\section{Methodology}

\noindent\textbf{Problem Formulation.} In the domain of sketch-based strand generation, we aim to develop a model capable of effectively converting sketch images into realistic hair strands. Let $\mathcal{I}_S$ denote the distribution of sketch images, and $\mathcal{S}$ represent the distribution of hair strands. Our task is to learn a mapping function $\psi: \mathcal{I}_S \rightarrow \mathcal{S}$. The core challenge lies in modeling diverse patterns of sketches while ensuring the generated strands exhibit realistic interactions and fine-grained details.

\noindent\textbf{Overview.} We propose a novel framework for sketch-based strand generation, as illustrated in Fig.~\ref{fig:strand_model}. Our approach consists of two main components: a learnable strand upsampling strategy and a multi-scale adaptive conditioning mechanism. The upsampling strategy leverages the concept of “next-scale prediction” to encode strands into multiple latent spaces, facilitating the gradual construction of full strands. The adaptive conditioning mechanism employs learnable visual tokens to handle the diverse patterns of sketch images, ensuring consistency across various levels of granularity.

\subsection{Learnable Strand Upsampling Strategy}
The core of our framework lies in the learnable strand upsampling strategy, which addresses the limitations of traditional fixed upsampling. This strategy involves encoding 3D strands into multiple latent spaces, each representing a guiding scale with distinct detail levels. The smallest guiding scale captures basic guide strands, while subsequent scales progressively add more strands, culminating in a complete hairstyle.

\subsubsection{Multi-scale strand encoding. \label{sec:vae}} 

To encode and decompose 3D strand data into several latent spaces denoting different scales, we propose utilizing a hierarchical structure for the autoencoder. Specifically, for strands $S\in\mathbb{R}^{N\times P\times3}$, where $N$ denotes the number of strands, $P$ denotes the number of 3D points in each strand, we first follow HAAR~\cite{sklyarova2024text} to employ a pre-trained strand VAE $\varepsilon$ to encode $S$ into the strand-level latent space to yield $\hat{S}\in\mathbb{R}^{N\times64}$. This is then converted to a hair map $H\in \mathbb{R}^{128\times128\times64}$ via fixed mapping for each scalp position. $H$ is further decomposed into a map set $\{H_k\}_{k=1}^K$ corresponding to $K$ different guiding scales, via max pooling with a kernel size of $2^{K-k}$ applied to $H$ for the $k$-th scale. Intuitively, $H_k$ with the smaller guiding scale (i.e., smaller $k$) represents the hairstyle more \textit{abstractly} with fewer strands, generally describing the coarse style of the current sample. As $k$ increases, the spatial dimension of $H_k$ grows, i.e., more strands depicting the details are added.

While directly utilizing $\{H_k\}$ for strands generation is straightforward, we find this set suffers from two primary information redundancies: (1) Obviously, the guiding strands from ascendant scales are inherently contained in subsequent scales. (2) Guiding strands can also provide contextual information for neighboring strands. To address these issues, we further process $\{H_k\}$ into a new set $\{\hat{H}_k\}$ to reduce the redundancy as follows:
\begin{equation}
    \hat{H}_k = \begin{cases}
    H_1, k = 1, \\
    H_k - \mathrm{tile}({H}_{k-1}), k\in{\{2,\cdots,K \}}
    \end{cases}
\end{equation}
where $\mathrm{tile}$ denotes spatially tiling each pixel into a $2\times2$ grid. Thus, latent maps corresponding to each scale, except for the smallest one, solely contain residual information relevant to the specific scale. To further facilitate the generation process, we train a multi-scale latent VAE set $\{\varepsilon^L_k\}_{k=1}^K$ respectively for each scale, yielding in the latent embedding set $\{\hat{h}_k\}_{k=1}^K$. 

\subsubsection{Generation by Learning to Upsample  \label{sec:learnable_upsampling}}

To effectively learn the coarse-to-fine procedure embedded in the sequence of latent embeddings $\{\latentemb\}$, we adopt a simple yet effective approach inspired by the next-scale prediction methodology proposed in VAR~\cite{tian2024visual}. This involves constructing a scale-wise autoregressive generation model to capture the hierarchical nature of the data. For each guiding scale $k$, we model the conditional distribution of the latent embeddings $\latentemb$ given embeddings from all preceding scales, denoted as $p(\hat{h}_k|\hat{h}_1,\cdots,\hat{h}_{k-1})$. To achieve this, we employ a transformer architecture augmented with diffusion heads, inspired by MAR~\cite{li2024autoregressive}.

The generation model is structured as an encoder-decoder framework, where both components utilize transformer backbones. During the training phase, unmasked strand tokens $\hat{h}_k^{unmask}$ are randomly selected, whereas during inference, the unmasked tokens represent those generated in the previous iterations. These tokens are then processed by the mask encoder $\mathcal{E}^{G}$, which outputs the encoded unmasked tokens that are merged with the mask tokens. This combined embedding sequence is subsequently used by the mask decoder $\mathcal{D}^G$ to predict a latent conditioning embedding sequence $z_k$. Importantly, latent embeddings from all preceding scales $\prevlatentembset$ are incorporated into the input of both the mask encoder $\mathcal{E}^{G}$ and the mask decoder $\mathcal{D}^G$, ensuring that the model has access to necessary context. The final output is obtained by denoising random Gaussian noise via a MLP denoiser $\mathcal{D}^{MLP}$ with latent conditioning embedding $z_k$, following the DDPM~\cite{ho2020denoising} framework. We optimize with diffusion denoising loss, formulated as: 
\begin{equation}
    \mathcal{L}_{diff} = \|\epsilon_t-\mathcal{D}^{MLP}(\hat{h}_k, z_k, t)\|^2
\end{equation}
where $t$ denotes a randomly chosen diffusion timestep, and $\epsilon_t$ denotes the corresponding random noise.

Compared to a straightforward autoregressive model, our approach offers significant advantages by effectively utilizing continuous latent embeddings. This allows us to circumvent the common problems associated with discrete tokenization, such as information loss and reduced fidelity, which can adversely affect the quality of the generated output. 
Furthermore, each scale within our framework yields a decodable intermediate representation, corresponding to tangible hair geometry. This unique property allows users to inspect early-stage generation results and iteratively refine input sketches, in contrast to VAR models, which require generating the complete sequence before assessment or adjustment.
Moreover, unlike the diffusion model in HAAR, our structure provides flexible control based on the embeddings from preceding scales, enabling the model to adaptively capture the intricate relationships between neighboring strands, thereby addressing the limitations of fixed upsampling methods such as nearest neighbor or bilinear interpolation. By learning to predict the next scale, our model generates strands with realistic interactions and patterns, a capability crucial for high-quality strand generation, as it ensures that the generated strands are not only visually plausible but also exhibit the complex dependencies observed in real-world data.

\subsection{Multi-scale Adaptive Conditioning  \label{sec:conditioning_mechanism}}

To effectively condition the generation process on sketch image $\mathbf{I}_S$, a straightforward approach might involve using a fixed pretrained backbone, such as DINOv2~\cite{oquab2023dinov2}, to extract features $\hat{c}$ from $\mathbf{I}_S$, similar to existing text-based or image-based methods. These extracted features could then be integrated into the generation model through an attention mechanism. However, as elaborated in Sec.~\ref{sec:intro}, while sketch images are inherently more suitable as conditioning inputs compared to text or raw images, they present unique challenges due to their inherent variability. This variability arises from the differing skill levels, artistic styles, and interpretations of individual users, leading to a wide array of sketch patterns. Such diversity can complicate the conditioning process, as it introduces inconsistencies that a fixed feature extraction method may not adequately address. Moreover, within our multi-scale framework, utilizing sketches that do not align with the target scale can result in either missing critical details or introducing redundant information, both of which can adversely affect the model's performance.

We tackle these challenges by proposing a novel multi-scale adaptive conditioning mechanism that is crucial for effectively managing diverse sketch inputs. During training, when generating latent maps for the $k$-th guiding scale, we adapt DINOv2 by injecting learnable scale-specific tokens ${\theta^l_k}$ into the last third of its transformer layers. This adaptation enables the model to dynamically adjust to the specific characteristics of each scale. Specifically, for the $l$-th transformer layer, we append a set of learnable scale-specific tokens $\theta^l$ to the layer's original input $\tilde{c}^{l-1}$, which denotes the latent embedding of $\mathbf{I}_S$. The output of each self-attention layer is then modified as follows:
\begin{align}    
    \{Q,K,V\}^{'} &= W_{Q,K,V}(\left[\tilde{c}^{l-1}; \theta^l\right])
    \\
    \tilde{c}^{l} &= \mathrm{Attn}(Q^{'}, K^{'}, V^{'})
\end{align}
To incorporate scale-related knowledge into $\{\theta^l_k\}$, we introduce an optimization objective that aligns sketches of varying granularity with the target guiding scale. Specifically, for each strand data $S$, initially, we obtain its corresponding sketch images $\{\mathbf{I}_{S_k}\}$ for each guiding scale, as detailed in Sec.~\ref{sec:vae}. Subsequently, we derive the adapted embedding $\tilde{c}_{i}$, with $i$ randomly sampled from $\{1,\cdots,K\}$, and compare it to the original DINO embedding $\hat{c}_{k}$. The objective is defined as:
\begin{equation}
    \mathcal{L}_{align} = \sigma(\hat{c}_{k})\log \sigma(\tilde{c}_{i})
\end{equation}
where $\sigma$ denotes the normalization following the DINO loss~\cite{oquab2023dinov2}. 
By optimization with this objective, our multi-scale conditioning mechanism ensures the model can effectively leverage the diverse patterns inherent in sketches, irrespective of their complexity or stylistic variations. This adaptability is crucial for preserving the integrity of generated output, as it enables the model to accommodate the wide range of sketch inputs it may encounter. Consequently, during inference, our framework maintains consistency and robustness when conditioned on sketches with various levels of granularity, ensuring that the generated strands are both global-shape accurate and visually realistic.

Furthermore, since sketches can convey both coarse and fine details of a hairstyle, we propose a dual-level conditioning mechanism to enhance sketch embeddings. This mechanism captures comprehensive sketch information by integrating local and global cues. 
Specifically, local patch tokens derived from the adapted DINOv2 are concatenated with strand tokens and interact through attention layers in the generation model, facilitating detail refinement. This interaction allows the model to focus on the intricate features of the sketch, ensuring that fine-grained details are accurately represented. 
Meanwhile, the class token, which encapsulates global information, is directly added to all strand tokens to guide the global shape. This ensures that the model retains a coherent understanding of the hairstyle's general geometry. This dual-level approach ensures that the model captures both the fine-grained details and the overarching structure of the hairstyle, resulting in more accurate and realistic strand generation. Through the integration of both local and global information, our approach provides a comprehensive framework for strand generation, resilient to the inherent variability in sketch inputs.

\subsection{Training Details}

\noindent\textbf{Training data process.} We collected 3D strands from two public datasets: USC-HairSalon~\cite{hu2015single} (343 hairstyles) and CT2Hair~\cite{shen2023ct2hair} (10 hairstyles), and further supplemented the dataset with 24 self-collected hairstyles. All hairstyles are aligned to a common head model, and each strand is resampled to 100 points. A training set of 333 hairstyles was randomly designated, with the remainder forming the test set. In line with HAAR~\cite{sklyarova2024text}, the training set is augmented via realistic variations including squeezing, stretching, cutting, and curliness. Corresponding sketch images are rendered with a standard upper-body human model under adaptive camera positions, ensuring frontal views capture comprehensive information. The shoulders and neck of the human model are used as references for hair length and volume. Subsequently, multi-scale sketch images are derived using a pre-trained line art extractor~\cite{zhang2023adding}.

\noindent\textbf{Optimization scheme.} To train the proposed framework, which comprises the latent VAE $\varepsilon$, adapter tokens $\theta$, and generation model $\mathcal{E}^G, \mathcal{D}^G, \mathcal{D}^{MLP}$ for each scale $k$, we adopt a multiple-stage training strategy. First, all latent VAEs are trained from scratch via reconstructing loss. Then $\theta_k$ is trained together with $\mathcal{E}^G, \mathcal{D}^G$ by composing the diffusion loss $\mathcal{L}_{diff}$ and alignment loss $\mathcal{L}_{align}$. To ensure the adapted DINOv2 extracts scale-specific features, we alternate between its embeddings for sketches with a random guiding scale and the original embeddings corresponding to sketches with the target guiding scale. Moreover, to mitigate cumulative errors from the gradual upsampling process, a random Gaussian noise belonging to $T$-th ($T <50 $) diffusion step is injected into the previous latent embeddings $\prevlatentembset$ during generating larger-scale guiding strands. This simulates the cumulative errors, enhancing model robustness.

\begin{figure*}[t]
  \centering
  \includegraphics[width=0.9\linewidth]{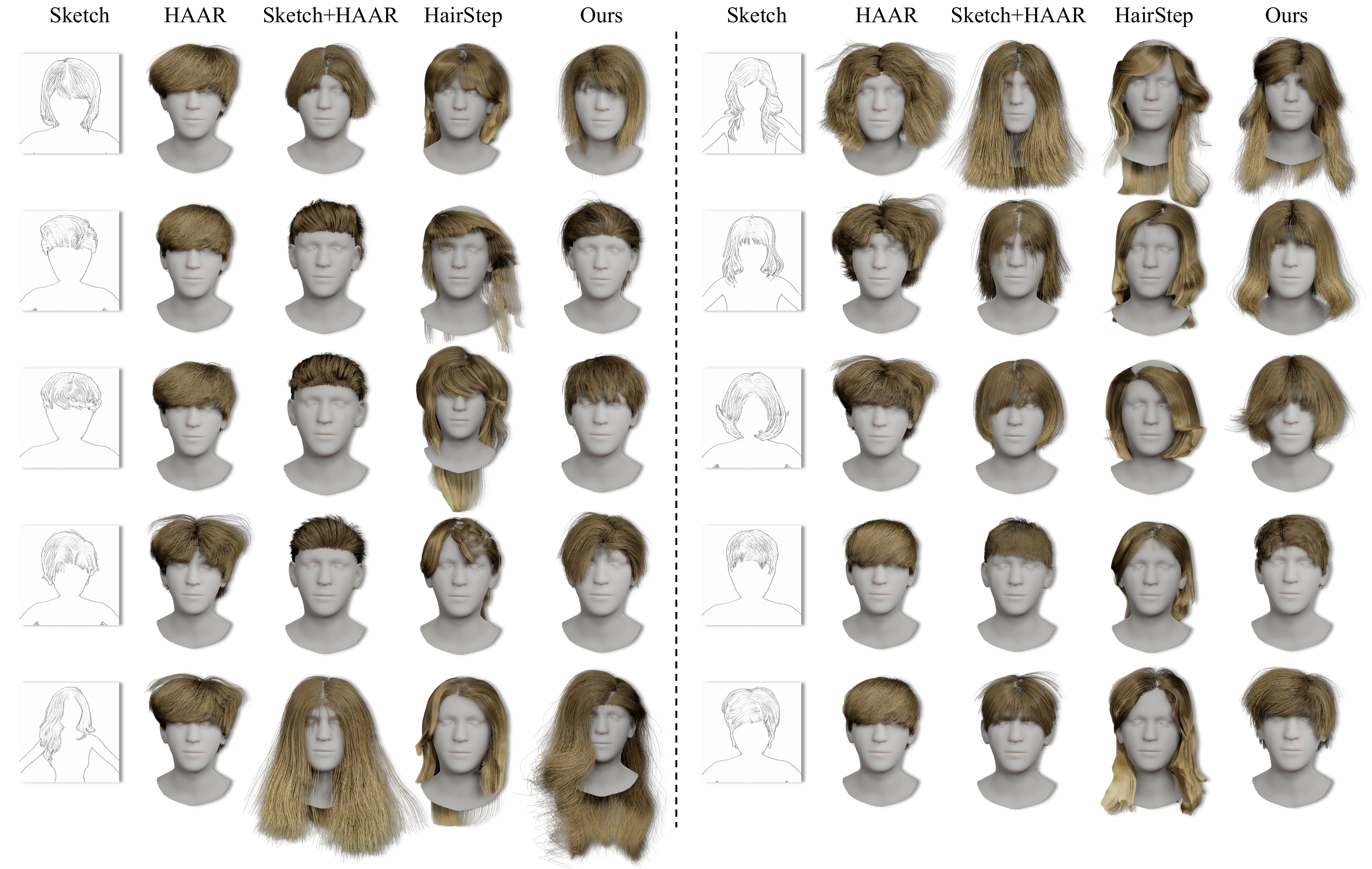}
  \vspace{-0.15in}
  \caption{
Qualitative comparison across HAAR~\cite{sklyarova2024text}, Sketch+HAAR, HairStep~\cite{zheng2023hairstep}, and our method. HAAR often ignores input conditions and defaults to short/medium styles. Sketch+HAAR improves accuracy via visual conditioning but still lacks detail fidelity. HairStep struggles with consistent geometry and produces artifacts such as incorrect back strands. In contrast, our method achieves high fidelity and precise control over diverse inputs.
}
  \label{fig:qualitative_comparison}
\end{figure*}

\section{Experiments \label{sec:experiment}}
\subsection{Implementation Details \label{sec:experiment-detail}}
We decompose each full hairstyle into three hair maps: \(H_1 \in \mathbb{R}^{32\times32\times64}\) (0.8k strands), \(H_2 \in \mathbb{R}^{64\times64\times64}\) (3k strands), and \(H_3 \in \mathbb{R}^{128\times128\times64}\) (12k strands). These maps are encoded using latent VAEs with compression rates of \(1/4\), \(1/16\), and \(1/16\), yielding latent embeddings of lengths 256, 256, and 1024, respectively. Besides, extreme outliers identified via the three-sigma rule in the hair maps are removed through local smoothing.
We adopt MAR-base as the backbone, featuring 12-layer transformer encoders/decoders and a hidden size of 768, following MAR's masking strategy. The training batch size is 256, with a learning rate of 1e-4 and 100 warm-up epochs. Classifier-Free Guidance~\cite{ho2022classifier}(CFG) is applied with a probability of 0.1 by randomly dropping the sketch condition.

\subsection{Quantitative Comparison}

Due to CFG, our method supports both unconditional and conditional hairstyle generation. Quantitative evaluations against state-of-the-art methods are conducted.

\noindent\textbf{Competitors.}
We compare our sketch-based strand generation model with two recent approaches:
(1) HAAR\cite{sklyarova2024text}, a text-guided strand generation model trained on LLaVA~\cite{liu2023visual} VQA answers. For fairness, we align with HAAR's conditioning by rendering front and side views for test hairstyles, processing with LLaVA, and using identical prompts and BLIP~\cite{li2022blip} features. To further evaluate the effect of sketch conditioning, we also re-implement a variant named “Sketch+HAAR” by replacing BLIP features with sketch-based DINOv2 features.
(2) HairStep\cite{zheng2023hairstep}, a single-view reconstruction method. We use SD3~\cite{esser2024scaling} to translate sketches into photorealistic images and retain only successful reconstructions.

\noindent\textbf{Unconditional Generation.}
Distribution quality is compared against HAAR~\cite{sklyarova2024text} using Chamfer Distance-based metrics: Minimum Matching Distance (MMD-CD) for sample quality, Coverage (COV-CD) for diversity, and 1-Nearest-Neighbor Accuracy (1-NNA) for overall distribution fit. As presented in Tab.~\ref{tab:unconditional}, our method demonstrates an improved capability in capturing the target hairstyle distribution, with lower MMD-CD (higher fidelity, closer to the references), and a higher COV-CD (greater diversity) than HAAR.

\begin{table}[t]
\caption{Comparison of Unconditional Generation}
\vspace{-0.1in}
\begin{tabular}{lccc}
\toprule
Method & MMD-CD $\downarrow$ & COV-CD(\%) $\uparrow$ & 1-NNA(\%) $\rightarrow$ 50\% \\
\midrule
HAAR~\cite{sklyarova2024text} & 0.0147 & 30.31 & 91.93 \\
Ours & \textbf{0.0090} & \textbf{35.17} & \textbf{88.95} \\
\bottomrule
\end{tabular}
\vspace{-0.1in}
\label{tab:unconditional}
\end{table}

\begin{table}[t]
\caption{Comparison of Conditional Generation}
\vspace{-0.1in}
\renewcommand{\tabcolsep}{1.5pt}
\begin{tabular}{lccccc}
\toprule
Method & PC-IoU(\%) $\uparrow$ & CD(\%)  $\downarrow$ & Hausdorff  $\downarrow$ & CLIP $\uparrow$ & LPIPS  $\downarrow$\\
\midrule
HAAR~\cite{sklyarova2024text} &53.83	&2.21	&0.1392	&0.9197	&0.2417\\
Sketch+HAAR &60.85	&1.06	&0.1093	&0.9411	&0.1804\\
HairStep~\cite{zheng2023hairstep} & 58.87	&1.86	&0.1514	&0.9433	&0.1968\\
Ours & \textbf{64.54} & \textbf{0.80} & \textbf{0.0959} & \textbf{0.9507} & \textbf{0.1483}\\
\bottomrule
\end{tabular}
\vspace{-0.1in}
\label{tab:conditional}
\end{table}

\begin{figure*}[t]
  \centering
  \includegraphics[width=0.9\linewidth]{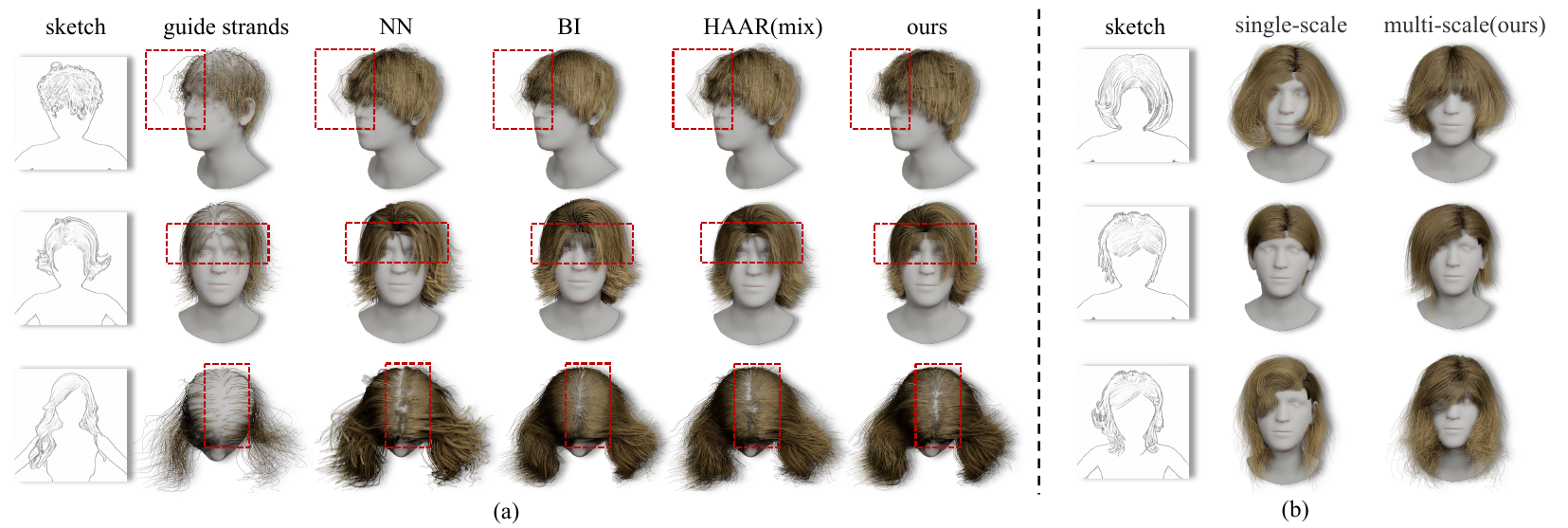}
  \caption{Ablation study on upsampling strategy. (\textbf{a}) Comparison to common alternatives (NN, BI, HAAR's fixed approach) shows their artifacts like clustering, over-smoothing, and inconsistencies (highlighted by red boxes). Our learnable approach better preserves fine details with realistic transitions. (\textbf{b}) Multi-scale (ours) vs. single-scale. The single-scale baseline fails to faithfully follow input sketches, while our progressive multi-scale strategy maintains superior conditional consistency.
  }
  \label{fig:ablation_upsampling}
\end{figure*}
\begin{figure}[t]
  \centering
  \includegraphics[width=\linewidth]{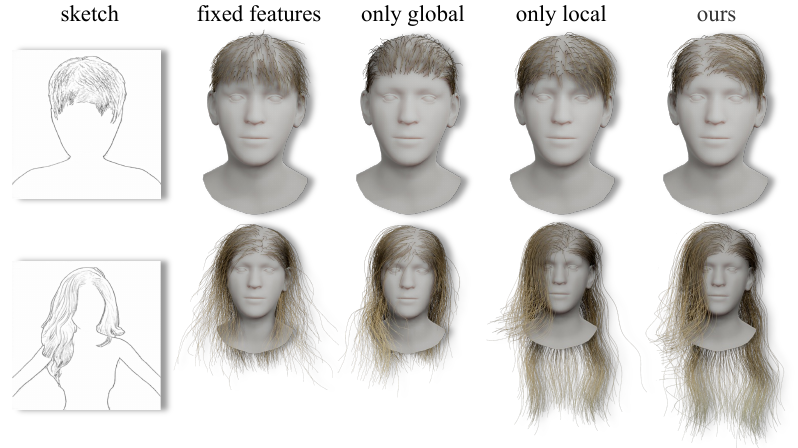}
  \caption{Ablation of the conditioning for initial strands: fixed features fail with dense sketches; global/local features alone compromise detail fidelity or structure control. Our method resolves density mismatch, preserving adherence.}
  \label{fig:ablation_global}
\end{figure}

\noindent\textbf{Conditional Generation.}
Conditional generation is evaluated via consistency with ground truth geometry and input conditions. Geometric fidelity relative to the ground truth is measured by Point Cloud IoU (PC-IoU), Chamfer Distance (CD), and Hausdorff Distance. 
Semantic alignment between front-view rendered outputs and input sketches is evaluated via CLIP Score~\cite{hessel2021clipscore} and LPIPS~\cite{zhang2018unreasonable}. 
As shown in Tab.\ref{tab:conditional}, compared with HAAR\cite{sklyarova2024text}, Sketch+HAAR, and HairStep~\cite{zheng2023hairstep}, our approach achieves consistent improvements across all metrics. 
Results demonstrate that our method generates hairstyles with higher geometric accuracy to the ground truth and stronger semantic adherence to conditioning sketches.

\subsection{Qualitative Comparison}
Fig.~\ref{fig:qualitative_comparison} shows qualitative comparisons of our method, HairStep~\cite{zheng2023hairstep}, HAAR~\cite{sklyarova2024text}, and re-implemented sketch-guided variant, Sketch+HAAR, across various hairstyles.
While HAAR uses text-based features for conditioning, its outputs often mismatch the target descriptions, tending to generate short or medium-length hairstyles irrespective of input prompts.
Sketch+HAAR better aligns with input conditions, demonstrating visual guidance's advantage, but still fails to capture fine-grained attributes like global length, silhouette, and bangs.
HairStep produces frontally similar hairstyles but suffers from inconsistencies, e.g., long back strands in short hairstyle reconstructions, leading to incoherent global shapes.
In contrast, our method consistently yields high-fidelity results closely matching the input across views, showcasing superior accuracy and controllability.

\subsection{Ablation Study}

To assess the effectiveness of our upsampling strategy and conditioning mechanism, we present ablation studies with intuitive qualitative results, including additional quantitative results in the supplementary material.

\noindent\textbf{Upsampling strategy.}
We evaluate the impact of different upsampling strategies, applying them to the same initial guide strands, with qualitative results in Fig.~\ref{fig:ablation_upsampling} (a). Our learnable upsampling approach is compared with common alternatives: Nearest Neighbor (NN) and Bilinear Interpolation (BI), which are standard methods in Blender~\cite{flavell2011beginning}, and HAAR~\cite{sklyarova2024text}'s fixed combination. While straightforward, NN interpolation tends to produce visible clustering artifacts unsuitable for realistic rendering. BI often results in excessive smoothing, particularly losing clear boundaries of hair partitions and bangs. HAAR's fixed combination yields inconsistent local patterns and struggles with diverse hair structures. 
In contrast, our learnable strategy, designed to be conditioned on both the guide strands and the input sketch, as detailed in Sec.~\ref{sec:learnable_upsampling}, effectively learns natural scale transitions and preserves fine-grained local hair details, delivering more realistic results.

We further compare our multi-scale approach to a baseline that directly generates full strands (single-scale), as show in Fig.~\ref{fig:ablation_upsampling}(b). 
This single-scale baseline employs the same fundamental architecture and parameters but directly generates approximately 12k full strands conditioned on the input sketch. 
Though producing complete hairstyles, it struggles to accurately adhere to sketch conditions. This is likely due to the increased difficulty of learning controllable conditional distribution for complex full hairstyles directly from limited data. Our multi-scale strategy proves more effective: first modeling the sparser guide strands distribution, then progressively refining details via the learnable upsampler, better preserving conditional consistency.

\noindent\textbf{Conditioning mechanism.}
We first evaluate our multi-scale adaptive conditioning mechanism, as elaborated in Sec.~\ref{sec:conditioning_mechanism}, against using fixed features, as shown in Fig.~\ref{fig:ablation_global}. When generating the initial (sparsest) guide strands from a relatively complex and dense input sketch, our adaptive approach  produces consistent strands, effectively handling the input-output density mismatch. In contrast, conditioning directly on fixed, pre-trained DINOv2~\cite{oquab2023dinov2} features leads to guide strands deviating from fine-grained sketch details under these challenging conditions.

Furthermore, we compare conditioning variants: global class token alone, local patch tokens alone, and our proposed dual-level fusion. The global token alone is insufficient to capture precise hairstyle details. Conversely, local tokens alone limit control over global structural attributes, such as overall length, parting style. Our dual-level design effectively leverages global context for structure and local features for detail, enabling more accurate and well-controlled hairstyle generation.

\subsection{Application and Discussion}
Our experiments confirmed the method's effectiveness, demonstrating strong controllability and high-quality generation across diverse hairstyles. 
This section further evaluates its practical utility and control capabilities by examining its response to sketches with varying densities, user edits, and hand-drawn styles.

\begin{figure}[htb]
  \centering
  \includegraphics[width=0.95\linewidth]{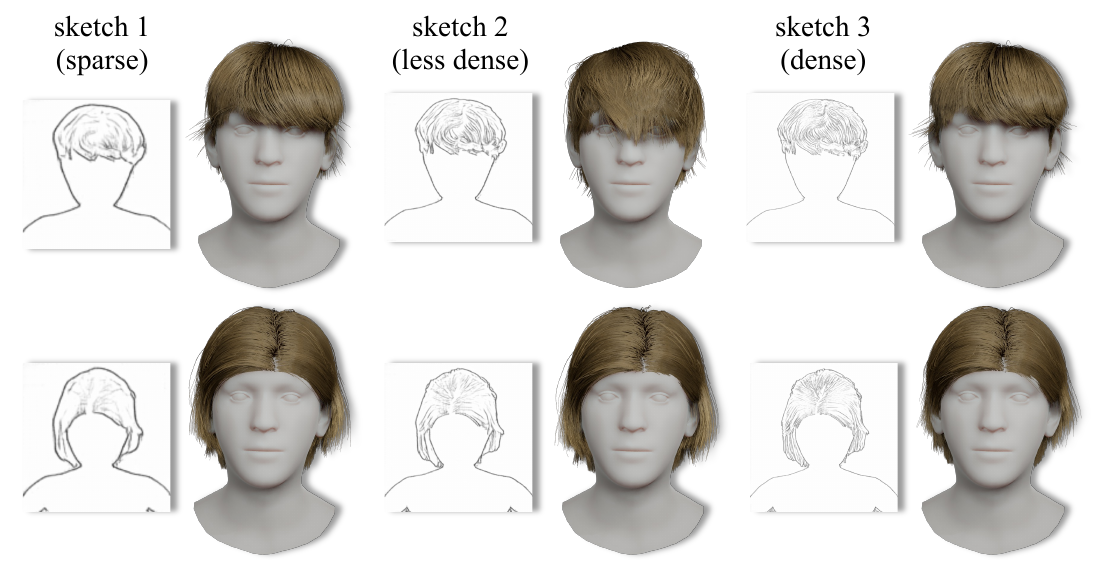}
  \caption{Adaptability to varying sketch densities. Consistent generation across the same hairstyle concept sketches with increasing densities (left to right)}
  \label{fig:app_density}
\end{figure}

\noindent\textbf{Adaptability to Sketch Density.}
Leveraging the multi-scale adaptive conditioning (Sec.~\ref{sec:conditioning_mechanism}), our method adapts to input sketches with varying granularity and density. Fig.~\ref{fig:app_density} demonstrates that for the same target hairstyle, input sketches with different densities yields consistent generation results aligned with the respective sketches, making the method accessible across user's professionalism.

\begin{figure}[htb]
  \centering
  \includegraphics[width=0.75\linewidth]{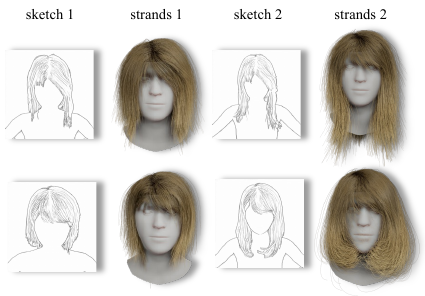}
  \vspace{-0.1in}
  \caption{Hairstyle modification via sketch editing. Changes sketch attributes ($\eg$ length/curliness) reflected in outputs.}
  \label{fig:app_control}
\end{figure}

\noindent\textbf{Can Users Control Hairstyles by Editing Sketches?}
Sketches inherently offer finer control and easier editability compared to text descriptions or raw images, as elaborated in Sec.~\ref{sec:intro}. Experiments with modified sketches, as in Fig.~\ref{fig:app_control}, demonstrate that users can modify attributes like hair length or curliness by simply sketching edits, highlighting the intuitive controllability of our sketch-based approach.

\begin{figure}[htb]
  \centering
  \includegraphics[width=0.95\linewidth]{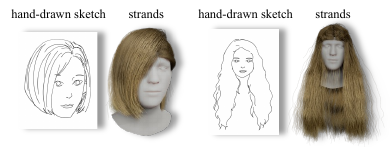}
  \caption{Generalization to hand-drawn sketches.}
  \label{fig:app_handdrawn}
\end{figure}

\noindent\textbf{Does the Method Generalize to Hand-drawn Sketches?} 
Although trained on sketches extracted from rendered images, our method exhibits generalization capacity to hand-drawn inputs. As shown in Fig.~\ref{fig:app_handdrawn}, it focuses on the hair region within the sketch and generates corresponding hairstyles regardless of whether the sketch includes facial features or body elements. While overall hairstyle generally conforms well to the hand-drawn sketch, certain fine details, such as precise parting lines or intricate hairlines, present opportunities for improvement. We attribute this limitation primarily to the scarcity and acquisition challenges of diverse, high-quality strand-based 3D hair data. Visual gaps between hand-drawn inputs and rendered training sketches, compounded by limited examples, further impede fine-detail capture. Consequently, future work should prioritize improved data acquisition for diverse sketch styles (especially hand-drawn) and more complex hairstyle structures.

\section{Conclusion}

We present the first sketch-conditioned framework for generating realistic 3D hair strands, which address the limitations of text and raw image prompts in existing methods, balancing user-friendliness with precise geometric control. By leveraging sketch inputs, our approach bridges this gap, enabling intuitive yet detailed hairstyle specification through binarized strokes. Extensive experiments show significant improvements over text-guided HAAR and image-based HairStep, with multiple quantitative metrics showing superior geometric fidelity and semantic alignment. Qualitative evaluations further validate our method's ability to generate realistic strands that respect sketch contours while maintaining physical plausibility. Future work can be extended to explore supporting multi-view sketch inputs and dynamic strand motion synthesis.

\bibliographystyle{ACM-Reference-Format}
\bibliography{sample-base}

\newpage
\appendix

{
\newpage
   \twocolumn[
    \centering
    \Large
    \textbf{Supplementary Material}\\
    \vspace{1.0em}
   ] 
}

\section{Additional Implementation Details}

\subsection{Evaluation Metrics}
Since there is no established benchmark for 3D hairstyle generation, we adopt evaluation metrics commonly used in the broader 3D shape generation literature. 

For \textit{unconditional generation}, we follow HAAR~\cite{sklyarova2024text} and assess generation quality, diversity, and distribution alignment using three Chamfer Distance (CD)-based metrics: (1) \textbf{Minimum Matching Distance (MMD-CD)}, which measures the average CD between each generated shape and its closest reference; (2) \textbf{Coverage (COV-CD)}, which computes the fraction of reference shapes matched by generated ones; and (3) \textbf{1-Nearest-Neighbor Accuracy (1-NNA)}, which evaluates whether generated and reference distributions are distinguishable based on local structure. 

For \textit{conditional generation}, we measure both geometric and semantic consistency with the input sketch. On the geometric side, we compute \textbf{Point Cloud IoU (PC-IoU)}, \textbf{Chamfer Distance (CD)}, and \textbf{Hausdorff Distance (HD)}, which are widely used in 3D point cloud evaluation. For semantic alignment, we adopt \textbf{CLIP Score}~\cite{hessel2021clipscore} and \textbf{LPIPS}~\cite{zhang2018unreasonable}, which compare the perceptual similarity between rendered results and the input sketch or image condition.

Let \( X = \{x_i\}_{i=1}^{n} \subset \mathbb{R}^3 \) and \( Y = \{y_j\}_{j=1}^{m} \subset \mathbb{R}^3 \) denote the ground-truth and generated point sets, respectively. All distances are computed using squared Euclidean distance. The metrics are defined as follows:

\begin{align}
\mathrm{CD}(X, Y) &= \frac{1}{n} \sum_{x \in X} \min_{y \in Y} \|x - y\|^2 + \frac{1}{m} \sum_{y \in Y} \min_{x \in X} \|y - x\|^2, \\
\mathrm{MMD}_{\mathrm{CD}} &= \frac{1}{|Y|} \sum_{Y_i \in Y} \min_{X_j \in X} \mathrm{CD}(X_j, Y_i), \\
\mathrm{COV}_{\mathrm{CD}} &= \frac{|\{ \arg\min_{X_j \in X} \mathrm{CD}(X_j, Y_i) \mid Y_i \in Y \}|}{|X|}, \\
\mathrm{1\text{-}NNA}(X, Y) &= \frac{1}{|X| + |Y|} \sum_{p \in X \cup Y} \mathbb{I}[\text{domain}(p) = \text{domain}(NN(p))], \\
\mathrm{PC\text{-}IoU}(X, Y) &= \frac{|\mathcal{V}(X) \cap \mathcal{V}(Y)|}{|\mathcal{V}(X) \cup \mathcal{V}(Y)|}, \\
\mathrm{HD}(X, Y) &= \max \left\{ \max_{x \in X} \min_{y \in Y} \|x - y\|, \; \max_{y \in Y} \min_{x \in X} \|y - x\| \right\},
\end{align}
where \( \mathcal{V}(X) \) denotes the set of occupied voxels after voxelizing the point cloud \( X \), and \( \text{domain}(p) \in \{X, Y\} \) indicates whether the point set \( p \) is real or generated. \( NN(p) \) denotes the nearest neighbor of \( p \) in the union \( X \cup Y \setminus \{p\} \), and \( \mathbb{I}[\cdot] \) is the indicator function.

\subsection{Dataset Process}
To obtain sketch image corresponding to 3D hair strands, we first rendered a standard upper-body human model featuring various hairstyles. Adaptive frontal camera views were employed during rendering to ensure each hairstyle occupied a significant central portion of the image, thereby capturing comprehensive structural details. Rendered body parts, such as shoulders and neck, served as visual scale references for hair length and volume.

\begin{figure}[t]
    \centering
    \includegraphics[width=\linewidth]{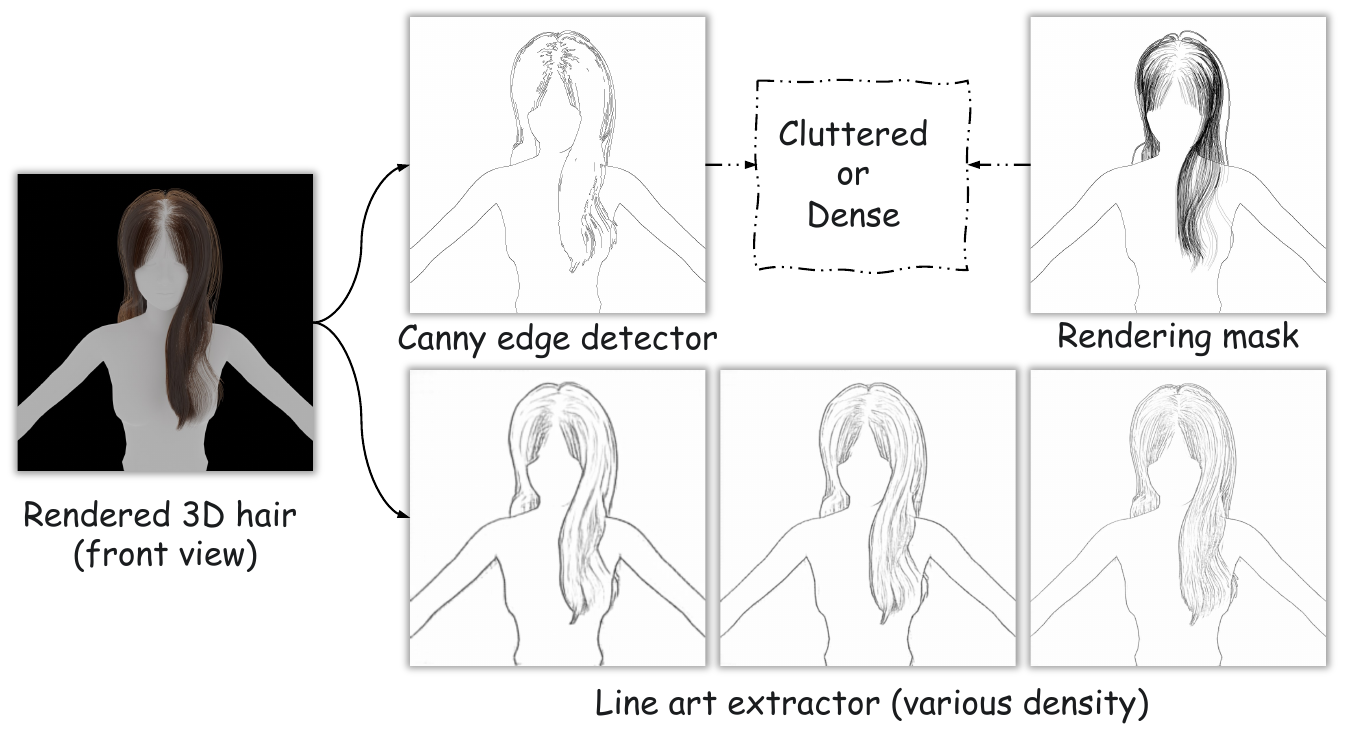}
    \caption{Sketch generation from rendered hair. Compared to Canny edges (often cluttered) or rendering masks (often too dense/lossy), the line art extractor produces sketches closer to hand-drawn styles with varying density levels.}
    \label{fig:supp_data}
    \vspace{-0.3in}
\end{figure}

Direct methods for generating sketch-like images from the renderings yielded suboptimal results. As illustrated in Fig.\ref{fig:supp_data}, edge detection using the Canny algorithm~\cite{canny1986computational} often produced cluttered contours with artifacts, deviating significantly from typical sketch styles. Directly utilizing rendering masks of the hair strands frequently resulted in overly dense outputs or the loss of critical structural information. Therefore, to generate sketches that more closely resemble actual hand-drawn inputs, we utilized a pre-trained line art extractor~\cite{zhang2023adding}. This extractor was used to generate sketch images at various density levels from the rendered views. Incorporating sketches with multiple densities into our training data aims to enhance the model's adaptability and generalization capabilities when faced with diverse real-world sketch inputs.

\begin{figure*}[t]
  \centering
  \includegraphics[width=0.9\linewidth]{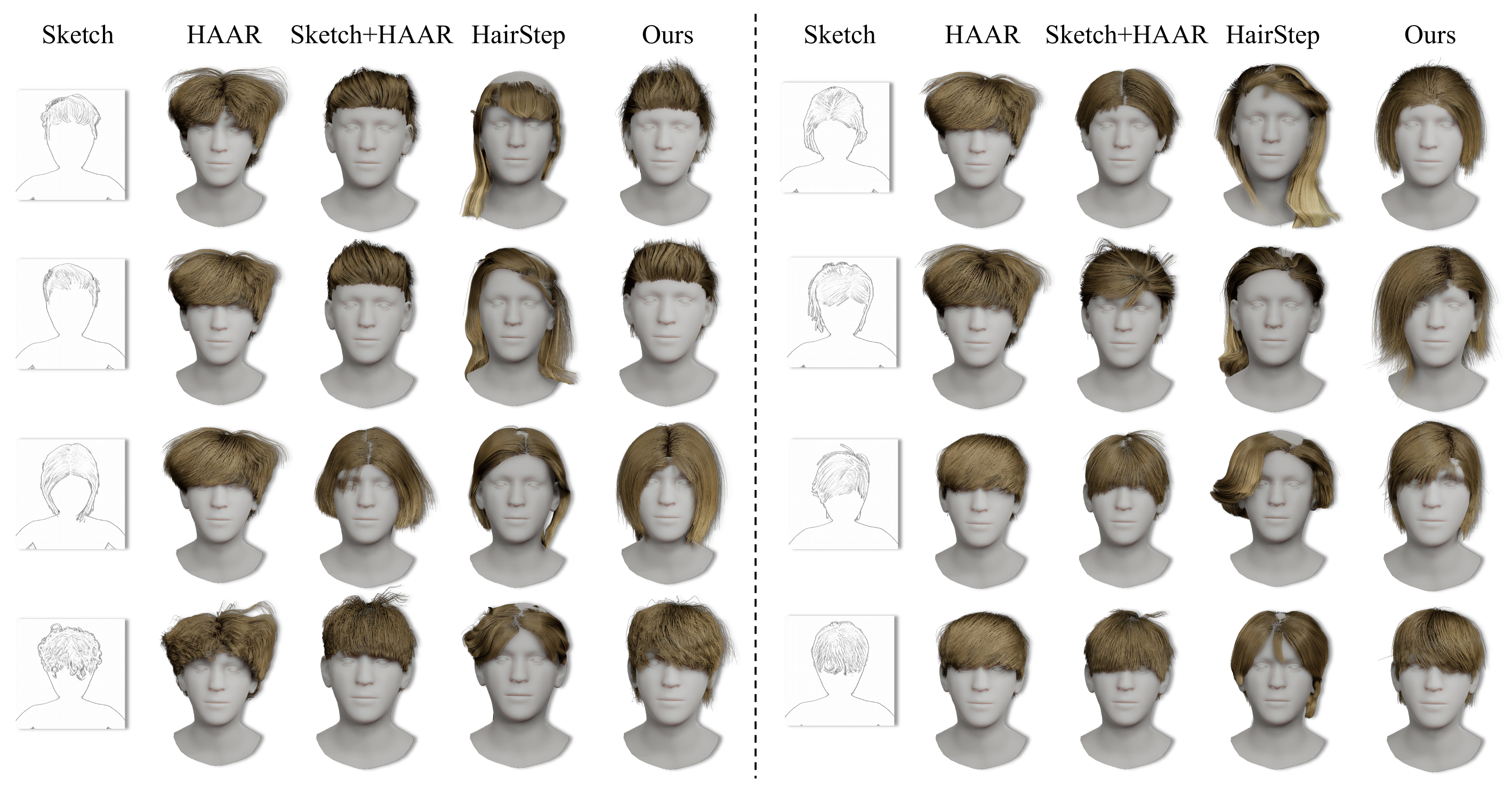}
  \caption{
Additional qualitative comparisons with HAAR~\cite{sklyarova2024text}, Sketch+HAAR, and HairStep~\cite{zheng2023hairstep}. HAAR often ignores input conditions, producing similar short hairstyles. Sketch+HAAR improves alignment but struggles with fine-grained details. HairStep shows geometric inconsistencies, such as erroneous long strands. In contrast, our method achieves high fidelity and accurately reflects diverse inputs.
}
  \label{fig:supp_compare}
\vspace{-0.1in}
\end{figure*}

\section{Additional Experiment Results}
\subsection{Qualitative Comparison}

To further substantiate the advantages of our proposed method, this section presents additional qualitative comparisons against HAAR~\cite{sklyarova2024text}, Sketch+HAAR, and HairStep~\cite{zheng2023hairstep} in Fig.\ref{fig:supp_compare}. Consistent with the findings discussed in the main paper, these supplementary results further illustrate the limitations of prior works. Specifically, HAAR tends to disregard input conditions, frequently generating similar short hairstyles with limited diversity. Sketch+HAAR improves alignment through visual conditioning but still struggles with fine-grained attributes such as length, silhouette, and bangs. HairStep exhibits inconsistent geometry, often producing artifacts such as erroneous long strands, even when reconstructing short hairstyles. 

In contrast, the results generated by our method across these diverse examples consistently exhibit high fidelity, realism, and precise adherence to the input conditions, reinforcing its superior controllability and effectiveness. This robust performance stems primarily from our framework's design, which synergistically employs a learnable multi-scale upsampling strategy via next-scale prediction and an adaptive conditioning mechanism with learnable visual tokens. The former enables the gradual construction of complex structures, while the latter ensures precise sketch adherence by capturing both global shape and local details.

Unlike GroomGen~\cite{zhou2023groomgen}, which reportedly relies on local strand interpolation, our approach explicitly integrates global context and local features throughout the multi-scale generation process. The strategy employed by GroomGen, focusing predominantly on local neighborhoods, might overlook broader hairstyle context or inadvertently smooth out fine-grained patterns. Since GroomGen is unavailable for direct comparison, these extensive experiments against available baselines serve to demonstrate the practical effectiveness and advantages of our approach. We believe our integration of multi-scale generation with adaptive conditioning offers valuable insights for the community pursuing controllable 3D content creation.

\vspace{-0.2in}
\subsection{Ablation Study}

To further validate the effectiveness of our upsampling strategy and conditioning mechanism, we provide additional quantitative results. Tab.\ref{tab:ablation_upsample} shows that our designed upsampling strategy significantly improves both geometric consistency and semantic similarity of the generated results. The proposed multi-scale design enables coarse-to-fine generation—modeling sparse guide strands at lower resolutions and refining structural details through the learnable upsampler. Tab.\ref{tab:ablation_scale} highlights the effect of varying the number of upsampling stages. We fix the maximum resolution and increase the number of scales $K$. As $K$ increases, the performance generally improves due to better coarse-to-fine modeling. However, overly large $K$ results in longer sequences with sparser information, which may degrade performance. Therefore, we avoid setting $K$ excessively high. Tab.\ref{tab:ablation_conditioning} demonstrates the benefit of our conditioning mechanism. It effectively integrates global structure and local details. These quantitative results align with our qualitative observations, further confirming the contribution of each component.

\begin{table}[t]
\centering
\caption{Ablation study on the upsampling strategy.}
\vspace{-0.1in}
\label{tab:ablation_upsample}
\small
\setlength{\tabcolsep}{4pt}
\begin{tabular}{lccccc}
\toprule
Method & PC-IoU(\%)$\uparrow$ & CD(\%)$\downarrow$ & Hausdorff$\downarrow$ & CLIP$\uparrow$ & LPIPS$\downarrow$ \\
\midrule
NN & 63.82 & 0.83 & 0.0985 & 0.9492 & 0.1621 \\
BI & 63.51 & 0.80 & 0.1035 & 0.9424 & 0.1690 \\
HAAR (mix) & 63.50 & 0.81 & 0.0990 & 0.9496 & 0.1584 \\
Ours & \textbf{64.54} & \textbf{0.80} & \textbf{0.0959} & \textbf{0.9507} & \textbf{0.1483} \\
\bottomrule
\end{tabular}
\vspace{-0.1in}
\end{table}

\begin{table}[t]
\centering
\caption{Ablation on the number of upsampling stages ($K$).}
\vspace{-0.1in}
\label{tab:ablation_scale}
\small
\setlength{\tabcolsep}{2pt}
\begin{tabular}{lcccccc}
\toprule
$K$ & Scale & PC-IoU(\%)$\uparrow$ & CD(\%)$\downarrow$ & Hausdorff$\downarrow$ & CLIP$\uparrow$ & LPIPS$\downarrow$ \\
\midrule
1 & 128 & 62.75 & 0.95 & 0.1038 & 0.9439 & 0.1705 \\
2 & 64$\rightarrow$128 & 63.05 & 0.84 & 0.1059 & 0.9474 & 0.1710 \\
3 (Ours) & 32$\rightarrow$64$\rightarrow$128 & \textbf{64.54} & \textbf{0.80} & \textbf{0.0959} & \textbf{0.9507} & \textbf{0.1483} \\
\bottomrule
\end{tabular}
\vspace{-0.1in}
\end{table}

\begin{table}[t]
\centering
\caption{Ablation on the conditioning mechanism.}
\vspace{-0.1in}
\label{tab:ablation_conditioning}
\small
\setlength{\tabcolsep}{4pt}
\begin{tabular}{lccccc}
\toprule
Method & PC-IoU(\%)$\uparrow$ & CD(\%)$\downarrow$ & Hausdorff$\downarrow$ & CLIP$\uparrow$ & LPIPS$\downarrow$ \\
\midrule
Only global & 60.86 & 0.95 & 0.1085 & 0.9457 & 0.1736 \\
Only local & 62.76 & \textbf{0.76} & 0.1077 & 0.9403 & 0.1755 \\
Fixed features & 63.01 & 0.89 & 0.0999 & 0.9499 & 0.1646 \\
Ours & \textbf{64.54} & 0.80 & \textbf{0.0959} & \textbf{0.9507} & \textbf{0.1483} \\
\bottomrule
\end{tabular}
\vspace{-0.1in}
\end{table}




\end{document}